\pdfoutput=1

\documentclass[11pt]{article}

\usepackage{acl}

\usepackage{times}
\usepackage{latexsym}

\usepackage[T1]{fontenc}

\usepackage[utf8]{inputenc}

\usepackage{microtype}

\title{The ITU Faroese Pairs Dataset}

\author{Leon Derczynski \\
  ITU Copenhagen \\
  Denmark 2300 \\
  \texttt{ld@itu.dk} \\\And
  Annika Solveig Hedegaard Isfeldt \\
  ITU Copenhagen \\
  Denmark 2300 \\ \AND
  Signhild Djurhuus \\
    ITU Copenhagen \\
  Denmark 2300 \\
  }

\begin{document}
\maketitle
\begin{abstract}
This article documents a dataset of sentence pairs between Faroese and Danish, produced at ITU Copenhagen. The data covers tranlsation from both source languages, and is intended for use as training data for machine translation systems in this language pair.
\end{abstract}

\section{Introduction}

Scandinavian languages are not particularly well resourced, coming from countries of 5-9 million people, the populations of which are generally reasonably well-educated and capable of switching to English when interfacing with technology, reducing demand on tools. While the national languages have some kind of NLP ecosystem around them, the other languages in the area are barely covered.

Under the system of placing resource availability for a given language on a class from 0-5 presented by~\citet{joshi-etal-2020-state}, the major national Scandinavian languages are Class~2. Nevertheless, the scarcity of native language tools reduces quality of life and technology adoption in the region.
Within each country there exist minority languages and dialects. The Danish Commonwealth, for example, has Class~1 S{\o}nderjysk and Class~0 Bornholmsk~\cite{derczynski2019bornholmsk} in Denmark, as well as three Inuit-Aleut languages on Greenland (Classes~0-1), and Faroese~\cite{knudsen2010language} in the Faroes. None of these languages have anywhere near the resources there are for Danish, which itself still struggles~\cite{kirkedal2019lacunae} despite modest tooling~\cite{derczynski2021danish,pauli2021danlp}.

There's especially little for Faroese which, as a Class 1 language whose strong cultural tradition is at risk~\cite{leonard2010faroese,haugen1979language}, isn't even covered by Google Translate. This limits interaction with speakers of other languages. On the other hand, the Faroe Islands have constructed their own alternative at {\small \url{https://www.faroeislandstranslate.com/}}, where one can enter an utterance in English and, if it's covered, will be responded to by a video of a Faroese person saying the translation in Faroese (if not, the translation request is sent out to native speakers, who can then submit a video translation).

To address this, we constructed a dataset of Faroese-Danish sentence pairs. The rationales, decisions, and methods used are described below in the format of a data statement for NLP~\cite{bender2018data}, which is a template for the paper.

\section{Executive Summary}

This is set of Faroese/Danish translated sentences. The goal is to help improve machine translation systems for Faroese. There are around 4000 sentence pairs in this dataset, using both Danish and Faroese source text. Translations are provided from L1 Faroese speakers who are highly proficient in Danish, living in Denmark.

\section{Curation Rationale}

The goal of this dataset is to provide as much impact as possible, given budget constraints, for machine translation for Faroese. Firstly, it is important to provide concepts and names native to the Faroe Islands, and so some of the source text has to be from and about that region. Secondly, a range of registers should be covered, so it is important to at least include both formal and informal text. Thirdly, the more language pairs that translations can be made with, the more versatile the data, so including Faroese translations of highly multilingual data has a good chance of being impactful.

To reach these targets, source translations in the ITU Faroese Pairs dataset are drawn from: a Faroese newspaper covering local news (Dimmalætting, {\small \url{https://www.dimma.fo/}}); the Europarl corpus of proceedings of the European Parliament, which is translated into 25+ languages~\cite{koehn2005europarl}; and Tatoeba, a multilingual dataset of informal conversational sentences ({\small \url{https://tatoeba.org/}}), which is released CC-BY 2.0 FR, using sentences that exist in Danish but not yet in Faroese.

\section{Documentation for Source Datasets}

The corpus relies on generating translations based on found data.

\paragraph{Dimmalætting} - a local newspaper; we only reuse sentences, none of which are over 250 words long, thus complying with the local copyright laws. The homepage is {\small \url{https://www.dimma.fo/}}. This news source is written in Faroese and primarily covers things of relevance to Faroese people. It is published weekly, based in the capital of the Faroes. Sentences from Dimmalætting used in this data were published between 2015--2019.

\paragraph{EuroParl} - a large corpus of European Parliament proceedings; this dataset uses a random selection of Danish sentence as sources to be translated into Faroese. See~\cite{koehn2005europarl} for details.

\paragraph{Tatoeba} - a collection of sentences and translations, mostly informal and conversational. Documentation is at {\small \url{https://en.wiki.tatoeba.org/articles/show/faq}}. Data is volunteer-submitted.

\section{Language Varieties}

In this case, the two languages are Faroese (bcp47 \texttt{fa}) and Danish (bcp47 \texttt{da-DK} and \texttt{da-FO}). The data covers the written and spoken styles; the Danish source texts are both from spoken language, of which Tatoeba is also informal.

\section{Speaker Demographic}

The speakers generating source texts vary depending on the source. For Dimmalætting, these are Faroese journalists operating mostly in the Faroes, writing the the late 2010s. For Europarl, the speakers are politicians from any EU country, working in the Europeal Parliament between from at least 2000 up to 2011. For Tatoeba, the speakers are proficient users of Danish from anywhere, who have engaged in the Tatoeba project.

\section{Annotator Demographic}

The annotators generating target translations are confined to Faroese native speakers in Copenhagen available for casual translation work. We ended up with a pool of two annotators, female, between 20 and 40, with a high level of education. For both, Faroese was their first language, they were highly proficient in Danish, and spoke English to at least a professional working level. This is mentioned because the annotation exercise and goals were described in English. The translations were produced between October 2019 and February 2020.

\section{Speech Situation and Text Characteristics}

The generated text is all written. Translators worked from home, adding translations to a spreadsheet sentence-by-sentence in blocks of 100. This was an asynchronous activity with no deadlines. Translators lived in Denmark during the work.

\section{Preprocessing and Data Formatting}

Source data was acquired as follows. For Dimalætting, some 200 articles were scraped from the site and the body text extracted, then tokenized using NLTK's tokenizer (there's no Faroese tokenizer, so given that any choice would probably make quite a few mistakes, a readily-accessible one seemed a reasonable choice). For Europarl, we use the Danish-English paired document, took 50\,000 sentences from this, and used the Danish version as a source. For Tatoeba, sentences that had Danish translations but not Faroese translations were used. 

The total accumulated text was then shuffled and used as sources for translation from either Faroese to Danish or Danish to Faroese. Sentences retain unique identifiers consisting of a prefix corresponding to their source, an ID number, and a three-letter language code suffix.

\section{Limitations}

Only a few genres of language are capture here. The text is only 4\,000 sentence pairs. There is no quality control on the output, even though it's produced by native speakers. There's no guarantee about the repetitiveness or diversity of the source data. The outputs cover only two native speakers' style. Each source sentence has only one target translation.

\section{Metadata}

\paragraph{License} The ITU Faroese Pairs dataset is licensed CC-BY 4.0, {\small \url{https://creativecommons.org/licenses/by/4.0/}}.
\paragraph{Annotation guidelines} Annotators were asked to provide translations that favoured being fluent in the target language over being faithful to the source language.
\paragraph{Annotation process} Source text was sent out in blocks of 200 in spreadsheets, that were filled in online.
\paragraph{Dataset quality metrics} None.

\section{Disclosure and Ethical Review}
There was no IRB review because the host organisation has no organ or procedure for this. Annotators were compensated at the local legal rate the university was legally constrained to pay them, and this rate is not public or trivial to derive. Annotators gave consent for the data to be released after annotation. The input data were not filtered for harmful content, even though given the source text nature, this is low-risk.

\section{Access}
The ITU Faroese Pairs dataset is available from the author and also via Hugging Face Datasets~\cite{lhoest-etal-2021-datasets} at {\small \url{https://huggingface.co/datasets/strombergnlp/itu_faroese_danish}}.

\section*{Acknowledgements}
This work was part of the Nordic e-Infrastructure Collaboration under the Nordic Language Processing Laboratory (NLPL),  and was 100\% funded by the Department of Computer Science at the IT University of Copenhagen.

\bibliography{anthology,custom}
\bibliographystyle{acl_natbib}

\end{document}